\pdfoutput=1
\documentclass[11pt]{article}

\usepackage{emnlp2021}

\usepackage{times}
\usepackage{latexsym}
\usepackage[T1]{fontenc}
\usepackage[utf8]{inputenc}
\usepackage{microtype}
\usepackage{placeins}
\usepackage{amsmath}
\usepackage{amssymb}
\usepackage{tabularx}
\usepackage{amsfonts}
\usepackage{url}
\usepackage{algorithmic}
\usepackage{fontawesome5}
\usepackage{graphicx}
\usepackage{textcomp}
\usepackage[table]{xcolor}
\usepackage{xcolor}
\usepackage{colortbl}
\usepackage{array}
\usepackage{booktabs}
\usepackage{multirow}
\newcommand{\bestmark}{$^{\ast}$}
\usepackage{tikz}
\usetikzlibrary{positioning, arrows.meta, calc}
\usepackage[most]{tcolorbox}

\title{
CrisisKD: Five-Stage Knowledge Distillation for Aspect-Level Sentiment and Emotion Analysis in Crisis Discourse
}

\author{
Marko Haralovi\'c$^{1,2}$ \quad
Onat Akca$^{1}$ \quad
Salih Eren Y\"ucet\"urk$^{1}$ \quad
Minsi Li$^{3}$ \quad
Mari\"et Theune$^{1}$ 
\\[6pt]
$^{1}$Faculty of Electrical Engineering, Mathematics and Computer Science,
University of Twente
\\
$^{2}$Faculty of Electrical Engineering and Computing,
University of Zagreb
\\
$^{3}$Faculty of Behavioural, Management and Social Sciences,
University of Twente
}

\begin{document}

\maketitle

\newcommand{\methodname}{CrisisKD}

\begin{abstract}
Identifying the target of emotional words or phrases in crisis situations, especially health-related ones, is important for understanding public concerns across cultural and linguistic contexts. We propose \methodname, a five-stage teacher--student knowledge distillation framework for aspect-level sentiment and emotion analysis on unannotated social media data. A teacher LLM generates aspect-level labels and reasoning traces that supervise a smaller student model across aspect extraction, syntactic parsing, opinion extraction, sentiment classification, and emotion classification. Using this framework, we construct and release a dataset containing 50,615 aspect-level labels, together with the annotation and fine-tuning scripts as open-source resources.\footnote{\url{https://github.com/MarkoHaralovic/CrisisKD}} The resulting student supports end-to-end ABSA and emotion detection at substantially lower inference cost than the teacher. On a manually annotated 500-tweet gold set, the 5-task Qwen2.5-7B student improves over the untuned model by 7.9 F1 points on aspect extraction, 17.0 points on emotion accuracy, and 6.5 points on sentiment accuracy. On the external ABEA benchmark, CrisisKD improves the same-model Qwen2.5-7B ICL baseline by 2.8 F1 points on ATE and 3.8 F1 points on joint ATE+AEC.
\end{abstract}

\section{Introduction}

During major health crises, such as the COVID-19 pandemic, public discourse becomes emotionally complex~\cite{abd2020covidtwitter}. Understanding this discourse requires more than identifying whether a statement is positive or negative; we must also identify the specific topic, or ``aspect,'' targeted by the emotion. For example, ``fear'' directed at the virus may imply willingness to cooperate with health measures, whereas ``fear'' directed at a vaccine may indicate reduced trust in its safety. This distinction is important because aggregate sentiment measures can obscure differences in public response ~\cite{pontiki2014semeval}. The same emotion may therefore signal different intentions depending on its target.

Aspect-Based Sentiment Analysis (ABSA) identifies topics and the  sentiments expressed toward them, but standard approaches~\cite{zhang2018absa} have largely focused on product reviews and coarse sentiment polarities rather than the fine-grained emotions, such as fear, anger, and hope, that shape behavior in crisis contexts. In this paper, we jointly model aspect extraction, aspect-based sentiment analysis, and aspect-level emotion detection on unannotated COVID-19 tweets.

A major barrier to aspect-level analysis is the lack of large COVID-19 datasets manually annotated for aspect-based sentiment and emotion classification~\cite{banda2021covidtwitter}. Existing datasets typically focus on named entities or sentence-level sentiment rather than aspect-level emotion categories. This limitation is particularly acute outside English, where annotated crisis data is even scarcer.

To reduce dependence on manual annotation, we propose \methodname{}, a five-stage teacher--student knowledge distillation framework. A large teacher model automatically annotates existing COVID-19 data and provides supervision for smaller student models. These students are more suitable for lower-resource deployment, and quantized variants can run on consumer hardware.

\methodname{} is informed by prior reasoning-based distillation work, including Syn-Chain~\cite{fan2025synchain}, which shows that syntactic
parsing and opinion extraction can improve ABSA knowledge distillation. Unlike approaches that assume existing aspect-level annotations, \methodname{} operates on raw text and generates both labels and reasoning traces for five tasks: aspect extraction, syntactic parsing,
opinion extraction, sentiment classification, and emotion classification. The teacher outputs serve as supervision for the student, allowing it to learn both task labels and intermediate reasoning steps. Because the pipeline requires only raw text and a dependency parser, it is
architecturally portable to other crisis domains and languages supported by suitable parsers such as spaCy~\cite{honnibal2020spacy}, although such transfer remains to be empirically validated.
\subsection{Research Questions and Contributions}
\label{sec:rq_contrib}

Our work addresses three questions:
\begin{itemize}
    \item \textbf{RQ1}: Can LLM-generated synthetic annotations enable effective ABSA training on domain-specific data without human aspect-level annotations?
    \item \textbf{RQ2}: Do auxiliary syntactic-parsing and opinion-extraction tasks improve student performance on aspect extraction, sentiment classification, and emotion classification?
    \item \textbf{RQ3}: How do aspect-level emotions and sentiments differ in COVID-19 discourse, and what insights do they provide?
\end{itemize}

Our main contributions are:
(1) a multi-step reasoning framework with explicit aspect extraction and emotion classification, architecturally adaptable to other crisis domains and languages;
(2) an openly released COVID-19 ABSA dataset with emotion labels, annotation code, teacher prompts, and fine-tuned model weights; and
(3) a student model that supports both downstream crisis-discourse analysis and automatic annotation of unannotated data, evaluated on a manually annotated gold set of 500 tweets.
\section{Related Work}

\textbf{Weak Supervision and Automatic Labeling}
Interpreting event-specific tweets, such as COVID-19 discourse, is challenging because manually annotated data are scarce. Researchers therefore use \textbf{weak supervision}, where noisy or automatically generated labels provide training signals. Automatically generated
pseudo-labels have proved viable for aspect-based sentiment analysis, particularly when syntactic structure is incorporated~\cite{negi2024hybrid,ratner2020snorkel}.

\textbf{Syntactic Dependency in ABSA}
Identifying the precise target of a sentiment requires modeling the grammatical relations that connect opinions to their targets. Prior work shows that syntax helps disentangle complex sentiments in English and
Chinese COVID-19 discourse \cite{hou2025covidabsa}. Accordingly, \methodname{} includes syntactic parsing as an auxiliary task.

\textbf{Emotion Detection in Crisis Discourse}
Fine-grained emotional targets, such as fear and anxiety, are important for understanding public health compliance and response \cite{abadi2021anxious}. Previous work trained downstream models from lexicon-based emotion annotations \cite{salsabila2023aspectcovid}, but
fixed lexicons struggle with context-dependent phenomena such as sarcasm and implicit emotion. \methodname{} instead uses contextualized reasoning from a large language model.

\textbf{Reasoning-Based Knowledge Distillation}
Recent work moves beyond label prediction by transferring reasoning traces from a teacher model to a smaller student model \cite{fan2025synchain}. Such traces provide useful inductive bias for ABSA and can outperform label-only supervision. Building on this broader direction, \methodname{} provides a complete five-stage pipeline for
unannotated data, jointly covering aspect extraction, syntactic parsing, opinion extraction, sentiment classification, and emotion classification.

\textbf{Instruction Tuning}
Prompt formulation also influences ABSA performance.
PFInstruct~\cite{cabello2024simpleabsa} shows that NLP-specific task prefixes improve results across ABSA subtasks. We follow this principle in the design of the \methodname{} prompts.

\textbf{Cross-Lingual and Cross-Domain Portability}
Many ABSA and emotion systems remain restricted to English
\cite{wu2025mabsa,zhang2022absa}. Multilingual parsers such as spaCy~\cite{honnibal2020spacy} provide CoNLL-U representations for many languages, while multilingual COVID-19 datasets exist in Arabic, Spanish, and Chinese
\cite{imran2022tbcov,hou2025covidabsa}. Because \methodname{} requires only raw text, a dependency parser, and a multilingual teacher model such as Qwen2.5-32B~\cite{qwen2_5}, it is architecturally portable beyond
English, although such transfer remains to be empirically validated.

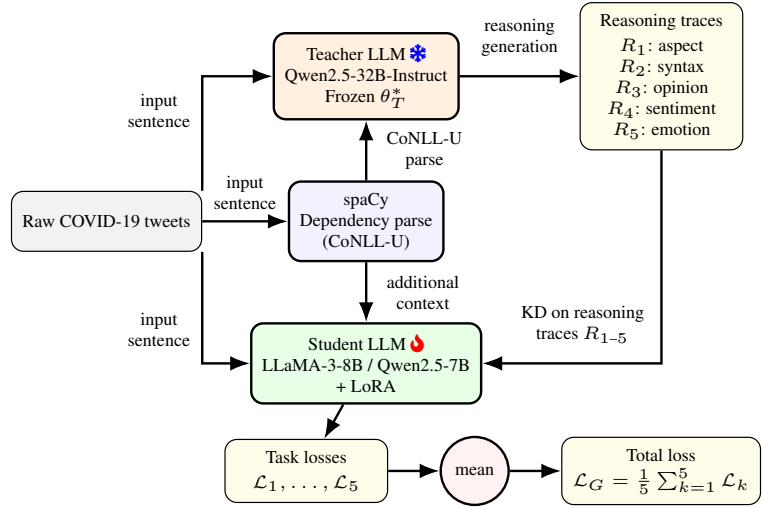
\begin{figure*}[t]
\centering
\newcommand{\figscale}{0.62}
\newcommand{\nodeMinW}{17mm}
\newcommand{\nodeMinH}{7mm}
\newcommand{\taskboxW}{19mm}
\newcommand{\hSepData}{10mm}
\newcommand{\hSepTeacher}{14mm}
\newcommand{\vSepTeacher}{7mm}
\newcommand{\vSepLoss}{4mm}
\newcommand{\nodeFontSize}{\tiny}

\resizebox{\figscale\linewidth}{!}{%
\begin{tikzpicture}[
    node distance=3mm and 6mm,
    module/.style={draw, rounded corners, thick, align=center,
                   fill=blue!5, minimum width=\nodeMinW,
                   minimum height=\nodeMinH, font=\nodeFontSize},
    moduleT/.style={module, fill=orange!12},
    moduleS/.style={module, fill=green!10},
    data/.style={draw, rounded corners, align=center,
                 fill=gray!10, minimum width=\nodeMinW,
                 minimum height=\nodeMinH, font=\nodeFontSize},
    arrow/.style={-Latex, thick},
    loss/.style={circle, draw, thick, fill=red!5,
                 minimum size=5mm, align=center, font=\nodeFontSize},
    taskbox/.style={draw, rounded corners, align=center,
                    fill=yellow!10, minimum width=\taskboxW,
                    font=\nodeFontSize}
]
\node[data] (data) {Raw COVID-19 tweets};
\node[module, right=\hSepData of data] (parser)
    {spaCy \\ Dependency parse \\ (CoNLL-U)};
\node[moduleT, above=\vSepTeacher of parser] (teacher)
    {Teacher LLM \textcolor{blue}{\faSnowflake} \\
     Qwen2.5-32B-Instruct \\
     Frozen $\theta_T^\ast$};
\node[moduleS, below=\vSepTeacher of parser] (student)
    {Student LLM \textcolor{red}{\faFire} \\
     LLaMA-3-8B / Qwen2.5-7B \\
     + LoRA};
\node[taskbox, right=\hSepTeacher of teacher] (traces)
    {Reasoning traces \\[0.5mm]
     $R_1$: aspect \\ $R_2$: syntax \\ $R_3$: opinion \\
     $R_4$: sentiment \\ $R_5$: emotion};

\draw[arrow] (data) --
  node[above, font=\nodeFontSize, align=center]{input\\sentence}
  (parser);
\draw[arrow] (data.north east) |-
  node[pos=0.35, left=0mm, font=\nodeFontSize, align=center]{input\\sentence}
  (teacher.west);
\draw[arrow] (data.south east) |-
  node[pos=0.35, left=0mm, font=\nodeFontSize, align=center]{input\\sentence}
  (student.west);
\draw[arrow] (parser) --
  node[midway, right=1mm, font=\nodeFontSize, align=center]{CoNLL-U\\parse}
  (teacher);
\draw[arrow] (parser) --
  node[midway, right=1mm, font=\nodeFontSize, align=center]{additional\\context}
  (student);
\draw[arrow] (teacher) --
  node[midway, above=1mm, font=\nodeFontSize, align=center]{reasoning\\generation}
  (traces);
\draw[arrow] (traces.south) |-
  node[pos=0.72, above=1mm, font=\nodeFontSize, align=center]
  {KD on reasoning\\traces $R_{1\text{--}5}$}
  (student.east);

\node[taskbox, below=\vSepLoss of student, xshift=-7mm] (losses)
    {Task losses \\[0.5mm] $\mathcal{L}_1,\dots,\mathcal{L}_5$};
\draw[arrow] (student) -- (losses);
\node[loss, right=6mm of losses] (avg) {mean};
\draw[arrow] (losses) -- (avg);
\node[taskbox, right=6mm of avg] (total)
    {Total loss \\[0.5mm]
     $\mathcal{L}_G=\frac{1}{5}\sum_{k=1}^{5}\mathcal{L}_k$};
\draw[arrow] (avg) -- (total);
\end{tikzpicture}%
}

\caption{System overview of the proposed teacher--student framework. Each input sentence is parsed by spaCy into CoNLL-U format and, together with the raw sentence, is fed to a frozen teacher LLM ($\theta_T^\ast$). The teacher produces multi-task reasoning traces $R_1$--$R_5$, which serve as knowledge-distillation targets for the student LLM (LLaMA-3-8B-Instruct or Qwen2.5-7B-Instruct, fine-tuned with LoRA). Task-specific generative losses $\mathcal{L}_1\ldots\mathcal{L}_5$ are averaged into $\mathcal{L}_G$.}
\label{fig:system_overview}
\end{figure*}

\section{Proposed Method}

\subsection{Method Overview}

To address the challenge of identifying emotional targets without large-scale manual annotation, we propose \methodname{}, a teacher--student knowledge distillation framework visualized in Fig.~\ref{fig:system_overview}.

\methodname{} uses the reasoning capabilities of a large language model (Teacher) to synthesize training data containing explicit reasoning traces that explain each label. These traces are then used to fine-tune a smaller, efficient Student model. This allows the Student to learn the reasoning required to identify why a sentiment or emotion is directed at a specific target while retaining the computational efficiency needed for large-scale analysis.

\subsection{Data Selection}

We selected COVID-19 tweet datasets that provide complete text for pseudo-annotation and knowledge distillation. Among available resources, including TweetsCOV19~\cite{dimitrov2020tweetscovid}, COVIDSenti~\cite{naseem2021covidsenti}, COVID19 NLP~\cite{covid19nlp}, TB-COV~\cite{imran2022tbcov}, and METS-CoV~\cite{ylab2022metscov}, many release only tweet IDs and therefore require Twitter/X API access. We consequently use \textbf{COVIDSenti}~\cite{naseem2021covidsenti}, with approximately 90{,}000 tweets labeled as positive, negative, or neutral, and \textbf{COVID19 NLP}~\cite{covid19nlp}, with 48{,}000 tweets labeled using five sentence-level sentiment categories. Neither dataset provides aspect-level sentiment or emotion labels.

We therefore pseudo-annotate both datasets with the teacher model, generating five reasoning traces ($R_{1-5}$) per sentence. These synthetic annotations are then used for knowledge distillation and student training. For cross-dataset evaluation, we additionally use the manually annotated \textbf{ABEA} dataset~\cite{zorenbohmer2026emograce}, which provides aspect and emotion annotations independent of our pseudo-annotation pipeline.

\textbf{Preprocessing} Tweets lacking personal opinion or emotion, including headline-style posts, links, and quoted news, were removed using rule-based filtering. The remaining tweets were normalized, stripped of emojis and URLs, and parsed with spaCy's~\cite{honnibal2020spacy} \texttt{en\_core\_web\_sm} model into CoNLL-U format, with Unicode normalization performed using \texttt{ftfy}~\cite{speer2019ftfy}. Full filtering criteria and preprocessing details are provided in Appendix~\ref{app:preprocessing}.

\subsection{Emotion Taxonomy}

For aspect-level emotion classification, we adopt a 14-category emotion taxonomy designed for crisis discourse, together with one additional \texttt{no\_emotion} fallback label for orphan annotations, failed parses,
or cases where the teacher does not assign a valid emotion. The SenWave dataset~\cite{yang2025senwave} suggests the following seven categories: \textit{optimistic}, \textit{thankful}, \textit{empathetic}, \textit{pessimistic}, \textit{anxious}, \textit{sad}, and \textit{annoyed}. We extend this set with seven additional categories derived from fine-grained emotion taxonomies used in prior work~\cite{demszky2020goemotions}: \textit{hopeful}, \textit{proud}, \textit{trustful}, \textit{satisfied}, \textit{scared}, \textit{angry}, and \textit{neutral}. We believe this taxonomy captures the details in emotional responses characteristic of pandemic discourse, distinguishing between related but distinct states such as \textit{anxious} versus \textit{scared}, or \textit{optimistic} versus \textit{hopeful}. The inclusion of no\_emotion allows the model to handle factual statements about aspects where no affective content is present.

Unlike sentence-level emotion detection, our task assigns exactly one emotion label per aspect, enabling fine-grained analysis of how different targets within the same tweet evoke distinct emotional responses. For instance, a tweet may express \textit{thankful} toward healthcare workers while simultaneously expressing \textit{anxious} toward vaccine side effects.

This design reflects a social-psychological perspective on crisis emotions: distinct states such as \textit{anxious} versus \textit{scared}, or \textit{optimistic} versus \textit{hopeful}, map to different behavioral intentions and compliance patterns in public health contexts~\cite{schroeder2016bayesact}. From the perspective of affect control theory, emotions are not merely internal states but socially and culturally shared sentiments that guide behavior~\cite{schroeder2016bayesact}. Fine-grained  aspect-level classification is therefore valuable beyond what coarse polarity models capture, particularly in multicultural crisis settings where the same emotional label may carry different behavioral implications across communities.

\subsection{The Reasoning Pipeline}

Our pipeline consists of five consecutive tasks. 

\textbf{Step 1: Aspect Extraction.} Building on the syntactic structure, the model identifies aspects: specific entities, topics, or targets of opinion. For example, in ``The vaccine rollout has been slow but the side effects are minimal,'' the aspects are ``vaccine rollout'' and ``side effects.'' The teacher model generates a reasoning trace $R_1$ explaining why each span constitutes an aspect (e.g., ``vaccine rollout'' is identified because it represents a concrete policy action that opinions are expressed about). This task enables our student model to perform automatic aspect annotation.

\textbf{Step 2: Syntactic Parsing.} Dependency parses are obtained via spaCy~\cite{honnibal2020spacy} and converted to CoNLL-U format, a tabular representation that explicitly encodes each word's part-of-speech, head word, and dependency relation. This structured format is easier for LLMs to process than raw dependency trees~\cite{fan2025synchain,matsuda2025dependency}. The teacher model generates a reasoning trace $R_2$ for syntactic parsing, learning to predict this structure from raw text. This serves two purposes: it provides syntactic features that inform the other tasks, and it acts as an auxiliary task that improves the student model's linguistic understanding.
    
\textbf{Step 3: Opinion Extraction.} For each identified aspect, the teacher model extracts the opinion terms expressing sentiment toward that aspect. In our example, ``slow'' expresses opinion toward ``vaccine rollout'' and ``minimal'' toward ``side effects.'' Reasoning trace $R_3$ explains each opinion-aspect connection through dependency relationships, providing significant performance gains~\cite{fan2025synchain}.

\textbf{Step 4: Sentiment Classification.} For each aspect-opinion pair, the model determines sentiment polarity (positive, negative, neutral). The reasoning trace $R_4$ explains why a particular sentiment label is appropriate given the opinion terms and broader context.

\textbf{Step 5: Emotion Classification.} For each aspect, the model identifies emotions expressed toward it (e.g., fear, anger, joy, trust, sadness). These categories are based on closed-set emotion taxonomies, which are relevant for crisis discourse. The reasoning trace $R_5$ explains the emotional content, which is more subtle than sentiment polarity.

Five reasoning traces were obtained from the teacher model Qwen2.5-32B-Instruct~\cite{qwen2_5}: $R_1$ for aspect extraction, $R_2$ for syntactic parsing, $R_3$ for opinion extraction, $R_4$ for sentiment classification, and $R_5$ for emotion extraction. The Qwen2.5-Instruct model family is reported to achieve state-of-the-art performance among open-weight models~\cite{qwen2_5}. Annotation of the corpus with Qwen2.5-32B-Instruct~\cite{qwen2_5} on an NVIDIA L40 (48\,GB) took approximately 7 days.
The student models are Llama-3-8B~\cite{llama3_1} and Qwen2.5-7B-Instruct~\cite{qwen2_5}, selected based on their strong distillation performance~\cite{fan2025synchain}. Both models are fine-tuned using a LoRA adapter~\cite{hu2022lora} via knowledge distillation from the teacher model's reasoning traces. We apply explicit teacher forcing between reasoning steps, i.e., step-level teacher annotations are provided as input rather than the student's predictions from previous step. We adopt LoRA as an effective substitute due to its efficiency and the computational cost of full fine-tuning~\cite{hu2022lora}.

\textbf{Loss definition} The unified training objective across all five tasks is formulated as a generative loss:
\begin{equation}
\mathcal{L}_G = -\frac{1}{N} \sum_{i=1}^{N} \sum_{t=1}^{T} \log P\big(g_{i,t} \mid \hat{g}_{i,<t}, C_i\big),
\end{equation}
where $N$ denotes the number of training samples, $T$ is the length of the output sequence for each sample, $g_{i,t}$ is the ground-truth token at position $t$ for sample $i$, $\hat{g}_{i,<t}$ is the previously generated token sequence up to position $t-1$, and $P(g_{i,t} \mid \hat{g}_{i,<t}, C_i)$ is the conditional probability of generating $g_{i,t}$ given the context $C_i$ (which encodes the input sentence, aspect, and task-specific prompt) and the prefix $\hat{g}_{i,<t}$~\cite{fan2025synchain}.

\textbf{Golden Set Evaluation}
The final fine-tuned model is evaluated against 500 manually annotated tweets derived from the COVID-Senti~\cite{naseem2021covidsenti} and COVID19-NLP~\cite{covid19nlp} datasets. Annotations followed a structured coding protocol. Inter-annotator agreement was assessed using span-level F1 scores for aspect extraction, and accuracy for sentiment and emotion labels on aligned aspect spans; disagreements were resolved through consensus. The finalized annotations constitute the gold standard used for evaluating model performance.

The main methodological contribution of \methodname{} is a
five-stage distillation pipeline that jointly generates aspect, syntax, opinion, sentiment, and emotion supervision from raw text. This enables the student to operate both as a downstream ABSA system and as an automatic annotator for previously unannotated data.

\subsection{Dataset Statistics}

Teacher annotations were aggregated over the COVID19NLP~\cite{covid19nlp} and COVIDSenti~\cite{naseem2021covidsenti} datasets. The resulting corpus contains 21{,}937 sentences and 50{,}615 aspect instances, corresponding to an average of 2.307 aspects per sentence. Neutral sentiment is the most
frequent label (51.87\%), followed by negative (36.13\%) and positive sentiment (12.01\%). The most frequent emotion labels are neutral (42.70\%), annoyed (20.73\%), and anxious (17.54\%). Full label distributions and data-quality statistics are reported in Appendix~\ref{app:dataset_statistics}.

\begin{table*}[t]
\centering
\caption{
Student--teacher fidelity across training scenarios.
We report precision, recall, F1, and accuracy for sentiment and emotion
classification, and precision, recall, and F1 for aspect extraction for each
model against teacher outputs. Bold values indicate the best performance
for a model, and \bestmark{} denotes the best performance overall.
}
\label{tab:distillation_results}

\setlength{\tabcolsep}{3.2pt}
\renewcommand{\arraystretch}{1.08}

\resizebox{\textwidth}{!}{%
\begin{tabular}{
    @{}
    l
    >{\raggedright\arraybackslash}p{4.0cm}
    ccc
    cccc
    cccc
    @{}
}
\toprule
\textbf{Model}
& \textbf{Scenario}
& \multicolumn{3}{c}{\textbf{Aspect Extraction}}
& \multicolumn{4}{c}{\textbf{Emotion Cls.}}
& \multicolumn{4}{c}{\textbf{Sentiment Cls.}} \\
\cmidrule(lr){3-5}
\cmidrule(lr){6-9}
\cmidrule(lr){10-13}
&
& \textbf{Prec.}
& \textbf{Rec.}
& \textbf{F1}
& \textbf{Prec.}
& \textbf{Rec.}
& \textbf{F1}
& \textbf{Acc.}
& \textbf{Prec.}
& \textbf{Rec.}
& \textbf{F1}
& \textbf{Acc.} \\
\midrule

\multirow{5}{*}{\textbf{Llama-3-8B-Instruct}}
& No fine-tuning
& 0.341 & 0.515 & 0.411
& 0.566 & 0.635 & 0.515 & 0.733
& 0.876 & 0.909 & 0.891 & 0.911 \\

& Fine-tune (3 tasks, w/o opinions and syntax)
& \textbf{0.654} & 0.643 & \textbf{0.648}
& 0.699 & 0.656 & 0.669 & 0.850
& \textbf{0.935}\bestmark
& \textbf{0.912}\bestmark
& \textbf{0.920}\bestmark
& \textbf{0.929} \\

& Fine-tune (4 tasks, w/o opinion extraction)
& 0.653 & 0.626 & 0.639
& 0.710 & 0.640 & 0.649 & 0.836
& 0.933 & 0.906 & 0.918 & 0.927 \\

& Fine-tune (4 tasks, w/o syntactic parsing)
& 0.629 & \textbf{0.646} & 0.638
& \textbf{0.724}\bestmark
& \textbf{0.657}
& \textbf{0.675}
& 0.852
& 0.933 & 0.899 & 0.915 & 0.926 \\

& Fine-tune (5 tasks)
& 0.649 & 0.622 & 0.635
& 0.698 & 0.640 & 0.638 & \textbf{0.858}
& 0.933 & 0.906 & 0.918 & 0.927 \\

\midrule

\multirow{5}{*}{\textbf{Qwen2.5-7B-Instruct}}
& No fine-tuning
& 0.466 & 0.461 & 0.463
& 0.561 & \textbf{0.710}\bestmark & 0.587 & 0.791
& 0.903 & 0.905 & 0.903 & 0.922 \\

& Fine-tune (3 tasks, w/o opinions and syntax)
& 0.638 & 0.658 & 0.648
& 0.699 & 0.682 & 0.670 & 0.852
& 0.930 & 0.900 & 0.915
& \textbf{0.929}\bestmark \\

& Fine-tune (4 tasks, w/o opinion extraction)
& \textbf{0.696}\bestmark & 0.728 & 0.712
& 0.694 & 0.689 & 0.663 & 0.910
& \textbf{0.933}
& \textbf{0.901}
& \textbf{0.916}
& 0.900 \\

& Fine-tune (4 tasks, w/o syntactic parsing)
& 0.665 & 0.723 & 0.693
& 0.707 & 0.673 & 0.667 & 0.920
& 0.929 & 0.900 & 0.914 & 0.920 \\

& Fine-tune (5 tasks)
& 0.681
& \textbf{0.747}\bestmark
& \textbf{0.712}\bestmark
& \textbf{0.710}
& 0.696
& \textbf{0.703}\bestmark
& \textbf{0.930}\bestmark
& 0.923 & 0.901 & 0.912 & 0.910 \\

\bottomrule
\end{tabular}%
}

\end{table*}

\section{Experiments}
We compare the performance of both student models on three tasks: aspect detection, sentiment classification, and emotion classification. Sentiment classification is a three-class problem (positive, negative, and neutral), while emotion classification uses 14 emotion categories and one additional \texttt{no\_emotion} fallback label.

We validate the benefit of training on all five pipeline stages by removing components of the pipeline and evaluating the resulting model performance. This allows us to assess whether improvements in syntactic parsing or opinion extraction lead to improved downstream performance on the classification tasks.

We additionally compare model performance before and after fine-tuning to validate the efficacy of our approach. Teacher-model outputs are used as references, allowing us to measure alignment between teacher and student predictions. The model was trained with teacher forcing, whereas validation was performed end-to-end by propagating student responses from earlier tasks, thereby reproducing inference-time evaluation. We report task-level precision, recall, and macro F1, as well as accuracy for sentiment and emotion classification.

We evaluate \methodname{} in three settings: student--teacher fidelity, in-domain performance on a manually annotated gold set, and cross-dataset transfer on the external ABEA benchmark.

\subsection{Training details}
We use LoRA, a parameter-efficient fine-tuning method, where only a small subset of weights (adapter) is trained~\cite{hu2022lora}. For fair comparison between the two student models, Llama-3-8B-Instruct and Qwen2.5-7B-Instruct, we use the same fine-tuning recipe. Specifically, we use a learning rate of $2\times10^{-5}$, mixed-precision training, batch size of 8, and gradient accumulation of 8, resulting in an effective batch size of 64. We use an adapter rank of 16 targeting projection layers. For Qwen2.5-7B-Instruct we train 40M parameters ($\sim$0.5\%), while for Llama-3-8B-Instruct we train 42M parameters ( $\sim$0.5\% of total parameters). The dataset is split into 80\% training and 20\% validation sets, and the exact splits are released with the code. All student models were trained for 500 steps on a single NVIDIA L40 (48\,GB), with training ranging 8-16 hours depending on the number of tasks in fine-tuning configuration.

\subsection{Student--Teacher Fidelity Evaluation}
Table~\ref{tab:distillation_results} reports results for fine-tuning Llama-3-8B-Instruct and Qwen2.5-7B-Instruct. We evaluate four settings: (i) no fine-tuning, (ii) fine-tuning on downstream tasks only (3 tasks: aspect extraction, aspect-based sentiment classification, and emotion classification), (iii) fine-tuning on all five reasoning steps (adding syntactic parsing and opinion extraction), and (iv) ablations that omit one of the two additional reasoning tasks (syntax or opinion extraction) to test whether these traces improve student performance.

It is important to note that knowledge distillation in this setup is not fully model-agnostic: the token distribution and reasoning style are typically more similar within a model family than across families. Consequently, Qwen2.5-7B-Instruct may learn more effectively from the in-family teacher Qwen2.5-32B-Instruct than Llama-3-8B-Instruct.

\textbf{Llama-3-8B-Instruct Results} From Table~\ref{tab:distillation_results}, we draw the following conclusions for Llama-3-8B-Instruct: adding reasoning traces for syntactic parsing and opinion extraction does not provide meaningful benefit. The model achieves the best overall performance when fine-tuned only on the three downstream tasks (aspect extraction, sentiment classification, and emotion classification), reaching the highest aspect extraction F1 score of $64.8\%$, the highest sentiment classification accuracy of $92.9\%$ across runs, and the second-highest emotion classification accuracy at $85.2\%$. Relative to the baseline, this corresponds to improvements of $+23.5$ pp in aspect extraction F1, $+1.8$ pp in sentiment classification accuracy, and $+11.7$ pp in emotion classification accuracy.

\textbf{Qwen2.5-7B-Instruct Results} For Qwen2.5-7B-Instruct, fine-tuning with reasoning traces yields substantial gains over the out-of-the-box baseline. The best overall configuration is obtained when fine-tuning on \textbf{all five tasks}, achieving the highest emotion classification accuracy ($93.0\%$) and the strongest aspect extraction performance (F1 of $71.2\%$). Removing either syntactic parsing or opinion extraction slightly reduces overall performance, suggesting that the additional reasoning steps are beneficial for the Qwen student model. Relative to the baseline, best model improves aspect extraction F1 by $+24.9$ pp, emotion classification accuracy by $+13.9$ pp, while sentiment classification, even though high $91.0\%$, drops by $-1.2$ pp compared to the baseline.

\textbf{Student Results Comparison} Comparing the best-performing configurations of both students, Llama (3 task fine-tuning) versus Qwen (5 task fine-tuning), we observe that the Qwen student consistently performs better on the downstream objectives. In particular, Qwen achieves higher aspect extraction performance (F1 score of $71.2\%$ vs.\ $64.8\%$) and significantly higher emotion classification accuracy ($93.0\%$ vs.\ $85.2\%$), indicating better alignment with the teacher outputs when distilling within the same model family.

\subsection{In-Domain Human Gold-Set Evaluation}
Inter-coder agreement on the double-annotated subsets indicates generally consistent annotations across annotators. Aspect extraction achieves a mean F1 score of $73.0\%$ across annotator pairs, while sentiment and emotion classification reach accuracies of $83.0\%$ and $79.0\%$, respectively.

Table~\ref{tab:gold_model_perf} reports results on the gold set of 500 tweets for the Qwen2.5-7B-Instruct baseline and our fine-tuned Qwen2.5-7B-Instruct students, trained on either 3 tasks or all 5 tasks in our pipeline, compared with the teacher model Qwen2.5-32B-Instruct. Fine-tuned models, as expected, outperform the baseline, with full fine-tuning across all 5 tasks showing better performance than 3-task fine-tuning. The performance of the fine-tuned model narrows the gap to the teacher model, demonstrating the effectiveness of our knowledge distillation framework.

\begin{table}[t]
\centering
\footnotesize
\caption{Performance on the manually annotated gold set. We report
accuracy and macro-F1 where applicable. Majority-class and uniform-random
baselines are included for the classification tasks.}
\label{tab:gold_model_perf}

\setlength{\tabcolsep}{3.2pt}
\renewcommand{\arraystretch}{1.05}

\begin{tabular}{@{}l l c c@{}}
\toprule
\textbf{Task} & \textbf{Model} & \textbf{Acc.} & \textbf{F1} \\
\midrule

\multirow{4}{*}{\shortstack[l]{Aspect\\Extraction}}
& Qwen2.5-7B Base      & -- & 0.549 \\
& Qwen2.5-7B FT (3)    & -- & 0.583 \\
& Qwen2.5-7B FT (5)    & -- & \textbf{0.628} \\
\cmidrule(l){2-4}
& \textit{Teacher: Qwen2.5-32B}
                        & -- & \textit{\textbf{0.709}} \\

\midrule

\multirow{6}{*}{\shortstack[l]{Emotion\\Classification}}
& Majority class       & \textit{0.551} & \textit{0.051} \\
& Uniform random       & \textit{0.305} & \textit{0.077} \\
\cmidrule(l){2-4}
& Qwen2.5-7B Base      & 0.508 & 0.569 \\
& Qwen2.5-7B FT (3)    & 0.599 & 0.562 \\
& Qwen2.5-7B FT (5)    & \textbf{0.678} & \textbf{0.570} \\
\cmidrule(l){2-4}
& \textit{Teacher: Qwen2.5-32B}
                        & \textit{\textbf{0.710}}
                        & \textit{\textbf{0.599}} \\

\midrule

\multirow{6}{*}{\shortstack[l]{Sentiment\\Classification}}
& Majority class       & \textit{0.537} & \textit{0.233} \\
& Uniform random       & \textit{0.423} & \textit{0.333} \\
\cmidrule(l){2-4}
& Qwen2.5-7B Base      & 0.658 & 0.692 \\
& Qwen2.5-7B FT (3)    & 0.713 & 0.696 \\
& Qwen2.5-7B FT (5)    & \textbf{0.723} & \textbf{0.696} \\
\cmidrule(l){2-4}
& \textit{Teacher: Qwen2.5-32B}
                        & \textit{\textbf{0.755}}
                        & \textit{\textbf{0.713}} \\

\bottomrule
\end{tabular}
\end{table}

\subsection{Cross-Dataset ABEA Evaluation}

\begin{table}[t]
\centering
\caption{
Performance on the ABEA test set. ATE denotes Aspect Term Extraction,
AEC denotes Aspect Emotion Classification, and ICL denotes in-context
learning. We compare our Qwen2.5-7B model against a same-size ICL
baseline and a larger Qwen3-14B ICL model as an upper-bound reference.
}
\label{tab:abea_model_perf}

\small
\renewcommand{\arraystretch}{1.15}
\setlength{\tabcolsep}{3pt}

\begin{tabular}{@{}l l c c c@{}}
\toprule
\textbf{Task} & \textbf{Model} & \textbf{Rec.} & \textbf{Prec.} & \textbf{F1} \\
\midrule

\multirow{3}{*}{ATE}
& \cellcolor{gray!12} Qwen3-14B ICL
& \cellcolor{gray!12} 0.468
& \cellcolor{gray!12} 0.427
& \cellcolor{gray!12} \textbf{0.447} \\

& Qwen2.5-7B ICL baseline
& 0.303 & 0.529 & 0.385 \\

& \textbf{Ours: Qwen2.5-7B}
& \textbf{0.359}
& 0.486
& \textbf{0.413} \\
\midrule

\multirow{3}{*}{\shortstack[l]{ATE +\\AEC}}
& \cellcolor{gray!12} Qwen3-14B ICL
& \cellcolor{gray!12} 0.358
& \cellcolor{gray!12} 0.327
& \cellcolor{gray!12} \textbf{0.342} \\

& Qwen2.5-7B ICL baseline
& 0.179 & 0.313 & 0.228 \\

& \textbf{Ours: Qwen2.5-7B}
& \textbf{0.231}
& \textbf{0.313}
& \textbf{0.266} \\
\bottomrule
\end{tabular}
\end{table}
\begin{table*}[t]
\centering
\small
\setlength{\tabcolsep}{6pt}
\renewcommand{\arraystretch}{1.16}
\begin{tabularx}{\textwidth}{@{}>{\bfseries}l X@{}}
\toprule
\textbf{Emotion} & \textbf{Top aspects (log-enrichment)} \\
\midrule

Proud
& \textit{doctor} (4.04), \textit{people} (1.69), \textit{vaccine} (1.48), \textit{quarantine} (1.47), \textit{shopping} (1.45) \\

Trustful
& \textit{doctor} (3.24), \textit{hand sanitizer} (2.53), \textit{World Health Organization} (2.52), \textit{mask} (2.38), \textit{government response} (1.89) \\

Optimistic
& \textit{shopping} (3.14), \textit{health} (1.75), \textit{cure} (1.74), \textit{food} (1.68), \textit{price} (1.66) \\

Thankful
& \textit{doctor} (3.13), \textit{health} (2.48), \textit{news} (2.26), \textit{grocery/supermarket/store} (1.98), \textit{shopping} (1.82) \\

Empathetic
& \textit{doctor} (3.06), \textit{people} (2.43), \textit{food} (2.15), \textit{health} (2.10), \textit{shopping} (1.78) \\

Hopeful
& \textit{cure} (3.04), \textit{vaccine} (2.48), \textit{news} (1.48), \textit{shopping} (1.30), \textit{world} (1.08) \\

Pessimistic
& \textit{economy} (3.03), \textit{world} (2.69), \textit{travel} (1.30), \textit{italy} (1.15), \textit{outbreak} (1.00) \\

Satisfied
& \textit{grocers response} (2.81), \textit{price} (2.52), \textit{shopping} (2.28), \textit{food} (2.01), \textit{flights} (1.73) \\

Sad
& \textit{doctor} (2.57), \textit{deaths} (1.77), \textit{quarantine} (1.19), \textit{people} (1.12), \textit{news} (0.92) \\

Angry
& \textit{racism} (2.50), \textit{trump} (1.93), \textit{public response} (1.84), \textit{government response} (1.63), \textit{government} (1.61) \\

No emotion
& \textit{vaccine} (2.39), \textit{mask} (2.16), \textit{doctor} (2.06), \textit{shopping} (1.67), \textit{government response} (0.80) \\

Annoyed
& \textit{media} (1.59), \textit{panic buying} (1.33), \textit{racism} (1.26), \textit{trump} (1.24), \textit{price} (1.03) \\

Anxious
& \textit{spread} (0.89), \textit{travel} (0.83), \textit{pandemic} (0.70), \textit{economy} (0.65), \textit{food} (0.40) \\

Neutral
& \textit{cruise ship} (0.58), \textit{death toll} (0.58), \textit{sars} (0.55), \textit{wuhan} (0.55), \textit{flights} (0.50) \\

Scared
& \textit{virus} (0.52), \textit{covid/coronavirus} (0.26), \textit{symptoms} (0.04), \textit{panic} (-0.10), \textit{travel} (-0.15) \\

\bottomrule
\end{tabularx}
\caption{Top aspects per emotion ranked by log-enrichment, defined as $\log\!\left(P(\text{emotion}\mid\text{aspect}) / P(\text{emotion})\right)$, over the top 50 aspects (minimum support $n \geq 200$). Positive values indicate that the emotion is more common for that aspect than in the dataset overall, negative values indicate underrepresentation, and 0 indicates baseline-level prevalence. Rows are ordered by the strongest top-ranked log-enrichment value in each emotion category.}
\label{tab:emotion_top_aspects_enrichment}
\end{table*}

Table~\ref{tab:abea_model_perf} evaluates cross-dataset transfer to the ABEA benchmark. Compared with the same Qwen2.5-7B model used with in-context learning (ICL) our fine-tuned model improves ATE F1 from 0.385 to 0.413 and joint ATE+AEC F1 from 0.228 to 0.266. This indicates that
our framework enables the 7B model to recover more
relevant aspect and aspect--emotion instances than prompting the same model with ICL alone, narrowing the gap to the larger Qwen3-14B model.

\subsection{Emotion--Aspect Associations via Enrichment Analysis}

To better characterize fine-grained emotional structure beyond aggregate distributions, we analyze the association between aspects and emotions using a log-enrichment metric (Table~\ref{tab:emotion_top_aspects_enrichment}). This measure points aspects for which a given emotion is disproportionately expressed relative to its overall frequency in the dataset.

The results reveal that different aspects exhibit distinctive emotional signatures. Positive, socially oriented emotions (e.g., \textit{proud}, \textit{trustful}, \textit{empathetic}) are strongly associated with human-centered aspects such as \textit{doctor} and \textit{people}, reflecting appreciation and collective solidarity. In contrast, negative emotions are concentrated around structurally different aspects: \textit{angry} and \textit{annoyed} are primarily linked to political and media-related aspects (e.g., \textit{racism}, \textit{trump}, \textit{media}), while \textit{pessimistic} and \textit{anxious} are associated with broader systemic concerns such as \textit{economy}, \textit{pandemic}, and \textit{travel}. Importantly, the same aspect appears across multiple emotion categories with different strengths. For example, \textit{doctor} is strongly enriched for \textit{proud}, \textit{trustful}, and \textit{empathetic}, while \textit{shopping} is associated with both \textit{optimistic} and \textit{satisfied}. This shows aspect-level sentiment cannot be fully captured by coarse polarity labels, as distinct emotional interpretations coexist within the same topic.

\section{Discussion}
With respect to the research questions introduced in Section~\ref{sec:rq_contrib}, we organize the following sections to show how our findings support each question.

\subsection{Effectiveness of Knowledge Distillation}
Our results show that student models achieve substantial improvements across aspect extraction, sentiment and emotion classification. For Llama-3-8B-Instruct, the best configuration scores a $+23.5$ pp improvement in aspect extraction F1, a $+1.8$ pp improvement in sentiment classification accuracy, and a $+11.7$ pp improvement in emotion classification accuracy over the non-fine-tuned baseline. Similarly, Qwen2.5-7B-Instruct achieves large gains in aspect extraction ($+24.9$ pp F1) and emotion classification ($+13.9$ pp accuracy), providing evidence for the effectiveness of knowledge distillation with reasoning traces.

To assess the contribution of syntactic parsing and opinion extraction, we train models using all five tasks and ablate omitting one or both of these auxiliary tasks. For Llama-3-8B-Instruct, adding syntactic parsing and opinion extraction does not lead to additional gains. We attribute this to cross-architecture distribution mismatch, where reasoning traces generated by Qwen2.5-32B-Instruct may not align well with the internal representations of Llama-3-8B-Instruct, or where limited training capacity is spent modeling the teacher reasoning style rather than improving downstream performance. 

In contrast, for the Qwen2.5-7B-Instruct student model, both syntactic parsing and opinion extraction provide measurable benefits. Fine-tuning on all five tasks results in the best overall performance, suggesting that the additional reasoning tasks are beneficial when teacher and student models belong to the same model family.

Overall, the results support claim that the proposed reasoning pipeline is effective. Exploring cross-architecture pairings and longer training schedules remains for future work.

\subsection{Emotion Patterns in COVID-19 Discourse}
Based on the teacher annotations, we observe that the majority of aspect-level sentiments are neutral (51.87\%), and that 42.70\% of aspect-level emotion labels are also neutral. At the sentence level, where annotations are available, we aimed to maintain similar label distributions.

It is possible for a sentence to exhibit an overall positive or negative valence while only the main aspect is assigned a corresponding sentiment/emotion label. Other, less central aspects that function as auxiliary nouns may be labeled as neutral because no explicit sentiment or emotional valence is expressed toward them. As a result, the distribution of neutral sentiment and emotion labels is skewed.

Another observation is the prevalence of negative sentiment labels, which occur approximately three times more frequently than positive labels, and negatively associated emotions, which appear about five times more often than positive ones. This pattern reflects the general negativity present in pandemic discourse~\cite{abadi2021anxious,lamsal2021covidsentiment}. In crisis situations, such as COVID-19 pandemic, it is therefore particularly important to identify the specific targets of negative emotions, which highlights the value of our aspect-level analysis.

\section{Conclusion}

We propose \methodname{}, a five-stage teacher--student knowledge distillation pipeline for aspect-level sentiment and emotion analysis on unannotated COVID-19 tweets. The pipeline combines aspect extraction, syntactic parsing, opinion extraction, sentiment classification, and emotion classification, using a 14-category emotion taxonomy with an additional \texttt{no\_emotion} fallback label. With Qwen2.5-32B as the teacher, we annotate 21,937 tweets with 50,615 aspect-level labels and fine-tune smaller student models, Llama-3-8B-Instruct and Qwen2.5-7B-Instruct, using LoRA. We evaluate the resulting models against both a manually annotated gold set of 500 tweets and the independently annotated ABEA benchmark. \methodname{} requires only raw text and a dependency parser, making it architecturally portable to other languages and crisis domains. Our contributions are:
\begin{itemize}
    \item We propose \methodname{}, a five-stage distillation pipeline for aspect-level sentiment and emotion analysis on unannotated data.

    \item We construct and release a COVID-19 ABSA dataset with 50,615 annotated aspects across 21,937 tweets, together with annotation code and fine-tuned model weights.

    \item We demonstrate effectiveness on both human-annotated and external benchmarks. On a manually annotated 500-tweet gold set, the 5-task Qwen2.5-7B student improves over the untuned model by 7.9 F1 points in aspect extraction, 17.0 points in emotion accuracy, and 6.5 points in sentiment accuracy. On the external ABEA benchmark, it improves over the same-model ICL baseline by 2.8 F1 points on ATE and 3.8 F1 points on joint ATE+AEC.
\end{itemize}
\begin{comment}
    \vspace{0.5em}
\noindent\textbf{Future Work.}
During manual evaluation of teacher annotations, we observed that teacher model produces semantically overlapping aspect predictions. To improve aspect granularity/precision, alternative prompting techniques, manual annotation of a small set of samples for in-context learning, and post processing check with a smaller model could be explored.

Second, future work should explore additional teacher--student architecture pairings and longer training schedules to explore cross-architecture knowledge transfer. Our results indicate that within same compute budget student models benefit more when distilling from a teacher within the same model family. While syntactic parsing and opinion extraction benefit the Qwen2.5-7B-Instruct student model, they do not consistently improve performance for Llama-3-8B-Instruct.

Finally, expanding manual evaluation to larger and more diverse datasets, as well as extending the framework to additional domains and languages, would further validate the generality of the proposed approach.

\end{comment}

\section*{Limitations}
The framework relies on teacher-generated annotations, so teacher errors and biases may propagate to the student models. Although the teacher generally produced well-formed outputs, its annotations should be treated
as noisy supervision rather than ground truth. In particular, it occasionally generated semantically overlapping aspect spans; because we prioritized recall over precision, we did not apply additional aspect-merging or span-refinement beyond prompt tuning. The 14-category emotion taxonomy may also miss subtle or culturally dependent expressions, while annotation with Qwen2.5-32B remains computationally expensive.

Nevertheless, the distilled students support lower-cost inference, including deployment on consumer hardware through quantization. The pipeline can reduce the cost of constructing domain-specific ABSA resources, and its reasoning traces support error analysis and system
refinement. Improved prompting, confidence-based filtering, or post-processing may further improve annotation precision. Although the method is architecturally portable to other domains and languages with suitable parsers, such transfer remains to be empirically validated.

\section*{Ethical Considerations}
This work studies aspect-based sentiment and emotion analysis on publicly available COVID-19 Twitter datasets. The datasets used in this research, COVIDSenti~\cite{naseem2021covidsenti} and COVID19NLP~\cite{covid19nlp}, consist of social media posts released for research purposes. No new human participant data were collected. Data was pseudo- and manually annotated, and annotations will be released to support future research. The datasets do not contain direct personally identifiable information, reducing privacy risks, and the analysis focuses on collective discourse patterns rather than individual users.

However, analyzing emotional content in social media raises ethical concerns. Systems that identify sentiment or emotion may be misused for surveillance, monitoring without consent, or manipulative targeting. While this work aims to support crisis communication and public opinion analysis, similar methods could negatively affect privacy or autonomy. Practitioners should carefully evaluate such risks.

The gold standard dataset was annotated by four volunteer annotators from the research team, without financial compensation. Annotators were trained using a coding scheme defining emotion and sentiment categories, examples, and rules for aspect spans, sarcasm, and edge cases. A calibration session was conducted before independent annotation. As annotators were team members performing an analytical task, this work did not undergo IRB review. No personal annotator data was collected, and participation was voluntary.

Bias and generalizability remain concerns. Social media data may not represent the full diversity of affected populations, and Twitter users are not demographically representative. Cultural differences also influence emotional expression. Additionally, the pseudo-annotation process relies on a large language model, which may introduce biases or errors. Thus, annotations should be interpreted as approximations of individuals' internal emotional states.

Although the proposed framework can be adapted to other languages and domains when raw text and linguistic analysis are available, the present study evaluates the approach only on COVID-19 Twitter data. The method assumes that aspect, sentiment, and emotion relations can be inferred from short-form text and syntactic structure; performance may degrade when these assumptions are violated (e.g., sarcasm, implicit context, or culturally dependent expressions). As a result, transfer to new settings remains to be validated with domain-specific data.

Finally, training large models has environmental implications. To mitigate this, the framework uses knowledge distillation and parameter-efficient fine-tuning (LoRA), reducing computational requirements, with only 0.5\% of model parameters trained.

To support transparency and reproducibility, we aim to open-source training and evaluation code, along with derived annotation artifacts where licensing permits, and future work should independently assess dataset and method bias to ensure ethical use of the proposed approach.

\section*{Acknowledgments} Computational resources were provided by the University of Twente High-Performance Computing infrastructure. Travel grants were provided by the Faculty of Electrical Engineering and Computing at the University of Zagreb.

\FloatBarrier

\bibliographystyle{acl_natbib}
\bibliography{custom}

@inproceedings{fan2025synchain,
  title     = {Aspect-Based Sentiment Analysis with Syntax-Opinion-Sentiment Reasoning Chain},
  author    = {Fan, Rui and Li, Shu and He, Tingting and Liu, Yu},
  booktitle = {Proceedings of the 31st International Conference on Computational Linguistics},
  month     = jan,
  year      = {2025},
  address   = {Abu Dhabi, UAE},
  publisher = {Association for Computational Linguistics},
  pages     = {3123--3137},
  url       = {https://aclanthology.org/2025.coling-main.210/}
}

@inproceedings{negi2024hybrid,
  title     = {A Hybrid Approach to Aspect Based Sentiment Analysis Using Transfer Learning},
  author    = {Negi, Gaurav and Sarkar, Rajdeep and Zayed, Omnia and Buitelaar, Paul},
  booktitle = {Proceedings of the 2024 Joint International Conference on Computational Linguistics, Language Resources and Evaluation (LREC-COLING 2024)},
  month     = may,
  year      = {2024},
  address   = {Torino, Italia},
  publisher = {ELRA and ICCL},
  pages     = {647--658},
  url       = {https://aclanthology.org/2024.lrec-main.56/}
}

@article{salsabila2023aspectcovid,
  title   = {Aspect-based Sentiment and Correlation-based Emotion Detection on Tweets for Understanding Public Opinion of Covid-19},
  author  = {Salsabila and Tyas, Salsabila Mazya Permataning and Romadhona, Yasinta and Purwitasari, Diana},
  journal = {Journal of Information Systems Engineering and Business Intelligence},
  volume  = {9},
  number  = {1},
  pages   = {84--94},
  year    = {2023},
  doi     = {10.20473/jisebi.9.1.84-94},
  url     = {https://doi.org/10.20473/jisebi.9.1.84-94}
}

@article{hou2025covidabsa,
  title   = {Aspect-based Sentiment Analysis for COVID-19: A Heterogeneous Graph Convolutional Network Approach},
  author  = {Hou, Linlin and Tu, Wenhui and Yu, Ting and Jiang, Ting and Bah, Mohamed and Xu, Zenghui and Zhang, Yu and Yang, Gaoming and Zhang, Ji},
  journal = {ACM Transactions on Asian and Low-Resource Language Information Processing},
  volume  = {24},
  number  = {6},
  pages   = {55:1--55:26},
  year    = {2025},
  doi     = {10.1145/3731758},
  url     = {https://doi.org/10.1145/3731758}
}

@inproceedings{cabello2024simpleabsa,
  title     = {It is Simple Sometimes: A Study On Improving Aspect-Based Sentiment Analysis Performance},
  author    = {Cabello, Laura and Akujuobi, Uchenna},
  booktitle = {Findings of the Association for Computational Linguistics: ACL 2024},
  month     = aug,
  year      = {2024},
  address   = {Bangkok, Thailand},
  publisher = {Association for Computational Linguistics},
  pages     = {6597--6610},
  doi       = {10.18653/v1/2024.findings-acl.394},
  url       = {https://aclanthology.org/2024.findings-acl.394/}
}

@inproceedings{dimitrov2020tweetscovid,
  title     = {{TweetsCOV19} -- A Knowledge Base of Semantically Annotated Tweets about the {COVID-19} Pandemic},
  author    = {Dimitrov, Dimitar and Baran, Erdal and Fafalios, Pavlos and Yu, Ran and Zhu, Xiaofei and Zloch, Matth{\"a}us and Dietze, Stefan},
  booktitle = {Proceedings of the 29th ACM International Conference on Information \& Knowledge Management},
  year      = {2020},
  pages     = {2991--2998},
  publisher = {Association for Computing Machinery},
  doi       = {10.1145/3340531.3412765},
  url       = {https://doi.org/10.1145/3340531.3412765}
}

@article{naseem2021covidsenti,
  title   = {{COVIDSenti}: A Large-Scale Benchmark Twitter Data Set for {COVID-19} Sentiment Analysis},
  author  = {Naseem, Usman and Razzak, Imran and Khushi, Matloob and Eklund, Peter W. and Kim, Jinman},
  journal = {IEEE Transactions on Computational Social Systems},
  year    = {2021},
  volume  = {8},
  number  = {4},
  pages   = {1003--1015},
  doi     = {10.1109/TCSS.2021.3051189},
  url     = {https://doi.org/10.1109/TCSS.2021.3051189}
}

@article{imran2022tbcov,
  title   = {{TBCOV}: Two Billion Multilingual {COVID-19} Tweets with Sentiment, Entity, Geo, and Gender Labels},
  author  = {Imran, Muhammad and Qazi, Umair and Ofli, Ferda},
  journal = {Data},
  volume  = {7},
  number  = {1},
  pages   = {8},
  year    = {2022},
  doi     = {10.3390/data7010008},
  url     = {https://doi.org/10.3390/data7010008}
}

@inproceedings{ylab2022metscov,
  title     = {{METS-CoV}: A Dataset of Medical Entity and Targeted Sentiment on {COVID-19} Related Tweets},
  author    = {Zhou, Peilin and Wang, Zeqiang and Chong, Dading and Guo, Zhijiang and Hua, Yining and Su, Zichang and Teng, Zhiyang and Wu, Jiageng and Yang, Jie},
  booktitle = {Advances in Neural Information Processing Systems},
  volume    = {35},
  pages     = {21916--21932},
  year      = {2022},
  publisher = {Curran Associates, Inc.},
  doi       = {10.52202/068431-1593},
  url       = {https://proceedings.neurips.cc/paper_files/paper/2022/hash/89a7ddfbc08b25ef8ff9029d7dd9e3d3-Abstract-Datasets_and_Benchmarks.html}
}

@inproceedings{hu2022lora,
  title     = {{LoRA}: Low-Rank Adaptation of Large Language Models},
  author    = {Hu, Edward J. and Shen, Yelong and Wallis, Phillip and Allen-Zhu, Zeyuan and Li, Yuanzhi and Wang, Shean and Wang, Lu and Chen, Weizhu},
  booktitle = {International Conference on Learning Representations},
  year      = {2022},
  url       = {https://openreview.net/forum?id=nZeVKeeFYf9}
}

@misc{covid19nlp,
  author       = {Miglani, Aman},
  title        = {Coronavirus tweets NLP - Text Classification},
  year         = {2020},
  howpublished = {Kaggle dataset},
  url          = {https://www.kaggle.com/datasets/datatattle/covid-19-nlp-text-classification},
  note         = {Accessed 2026-01-22}
}

@inproceedings{demszky2020goemotions,
  title     = {{G}o{E}motions: A Dataset of Fine-Grained Emotions},
  author    = {Demszky, Dorottya and Movshovitz-Attias, Dana and Ko, Jeongwoo and Cowen, Alan and Nemade, Gaurav and Ravi, Sujith},
  booktitle = {Proceedings of the 58th Annual Meeting of the Association for Computational Linguistics},
  year      = {2020},
  month     = jul,
  address   = {Online},
  publisher = {Association for Computational Linguistics},
  pages     = {4040--4054},
  doi       = {10.18653/v1/2020.acl-main.372},
  url       = {https://aclanthology.org/2020.acl-main.372/}
}

@misc{yang2025senwave,
  title         = {SenWave: A Fine-Grained Multi-Language Sentiment Analysis Dataset Sourced from COVID-19 Tweets},
  author        = {Yang, Qiang and Chen, Xiuying and Ma, Changsheng and Yin, Rui and Gao, Xin and Zhang, Xiangliang},
  year          = {2025},
  eprint        = {2510.08214},
  archivePrefix = {arXiv},
  primaryClass  = {cs.CL},
  doi           = {10.48550/arXiv.2510.08214},
  url           = {https://arxiv.org/abs/2510.08214}
}

@misc{qwen2_5,
  title = {{Qwen2.5} Technical Report},
  author = {
    Yang, An and
    Yang, Baosong and
    Zhang, Beichen and
    Hui, Binyuan and
    Zheng, Bo and
    Yu, Bowen and
    Li, Chengyuan and
    Liu, Dayiheng and
    Huang, Fei and
    Wei, Haoran and
    Lin, Huan and
    Yang, Jian and
    Tu, Jianhong and
    Zhang, Jianwei and
    Yang, Jianxin and
    Yang, Jiaxi and
    Zhou, Jingren and
    Lin, Junyang and
    Dang, Kai and
    Lu, Keming and
    Bao, Keqin and
    Yang, Kexin and
    Yu, Le and
    Li, Mei and
    Xue, Mingfeng and
    Zhang, Pei and
    Zhu, Qin and
    Men, Rui and
    Lin, Runji and
    Li, Tianhao and
    Tang, Tianyi and
    Xia, Tingyu and
    Ren, Xingzhang and
    Ren, Xuancheng and
    Fan, Yang and
    Su, Yang and
    Zhang, Yichang and
    Wan, Yu and
    Liu, Yuqiong and
    Cui, Zeyu and
    Zhang, Zhenru and
    Qiu, Zihan
  },
  year          = {2024},
  eprint        = {2412.15115},
  archivePrefix = {arXiv},
  primaryClass  = {cs.CL},
  doi           = {10.48550/arXiv.2412.15115},
  url           = {https://arxiv.org/abs/2412.15115}
}

@misc{llama3_1,
  title         = {The {Llama} 3 Herd of Models},
  author        = {Grattafiori, Aaron and Dubey, Abhimanyu and Jauhri, Abhinav and others},
  year          = {2024},
  eprint        = {2407.21783},
  archivePrefix = {arXiv},
  primaryClass  = {cs.AI},
  doi           = {10.48550/arXiv.2407.21783},
  url           = {https://arxiv.org/abs/2407.21783}
}

@article{abadi2021anxious,
  title   = {Anxious and Angry: Emotional Responses to the COVID-19 Threat},
  author  = {Abadi, David and Arnaldo, Irene and Fischer, Agneta},
  journal = {Frontiers in Psychology},
  year    = {2021},
  volume  = {12},
  pages   = {676116},
  doi     = {10.3389/fpsyg.2021.676116},
  url     = {https://doi.org/10.3389/fpsyg.2021.676116}
}

@misc{honnibal2020spacy,
  author       = {Honnibal, Matthew and Montani, Ines and Van Landeghem, Sofie and Boyd, Adriane},
  title        = {spaCy: Industrial-strength Natural Language Processing in Python},
  year         = {2020},
  howpublished = {Zenodo},
  doi          = {10.5281/zenodo.1212303},
  url          = {https://doi.org/10.5281/zenodo.1212303}
}

@misc{speer2019ftfy,
  author       = {Speer, Robyn},
  title        = {ftfy},
  year         = {2019},
  month        = mar,
  note         = {Version 5.5.1},
  howpublished = {Zenodo},
  doi          = {10.5281/zenodo.2591652},
  url          = {https://doi.org/10.5281/zenodo.2591652}
}

@article{abd2020covidtwitter,
  title   = {Top Concerns of Tweeters During the {COVID-19} Pandemic: Infoveillance Study},
  author  = {Abd-Alrazaq, Alaa and Alhuwail, Dari and Househ, Mowafa and Hamdi, Mounir and Shah, Zubair},
  journal = {Journal of Medical Internet Research},
  volume  = {22},
  number  = {4},
  pages   = {e19016},
  year    = {2020},
  doi     = {10.2196/19016},
  url     = {https://doi.org/10.2196/19016}
}

@inproceedings{pontiki2014semeval,
  title     = {{SemEval}-2014 Task 4: {Aspect Based Sentiment Analysis}},
  author    = {Pontiki, Maria and Galanis, Dimitris and Pavlopoulos, John and Papageorgiou, Harris and Androutsopoulos, Ion and Manandhar, Suresh},
  booktitle = {Proceedings of the 8th International Workshop on Semantic Evaluation ({SemEval} 2014)},
  year      = {2014},
  month     = aug,
  address   = {Dublin, Ireland},
  publisher = {Association for Computational Linguistics},
  pages     = {27--35},
  doi       = {10.3115/v1/S14-2004},
  url       = {https://aclanthology.org/S14-2004/}
}

@article{zhang2018absa,
  title   = {Deep learning for sentiment analysis: A survey},
  author  = {Zhang, Lei and Wang, Shuai and Liu, Bing},
  journal = {Wiley Interdisciplinary Reviews: Data Mining and Knowledge Discovery},
  volume  = {8},
  number  = {4},
  pages   = {e1253},
  year    = {2018},
  doi     = {10.1002/widm.1253},
  url     = {https://doi.org/10.1002/widm.1253}
}

@article{banda2021covidtwitter,
  title   = {A Large-Scale {COVID-19} {Twitter} Chatter Dataset for Open Scientific Research---An International Collaboration},
  author  = {Banda, Juan M. and Tekumalla, Ramya and Wang, Guanyu and Yu, Jingyuan and Liu, Tuo and Ding, Yuning and Artemova, Ekaterina and Tutubalina, Elena and Chowell, Gerardo},
  journal = {Epidemiologia},
  volume  = {2},
  number  = {3},
  pages   = {315--324},
  year    = {2021},
  doi     = {10.3390/epidemiologia2030024},
  url     = {https://doi.org/10.3390/epidemiologia2030024}
}

@article{ratner2020snorkel,
  title   = {Snorkel: Rapid Training Data Creation with Weak Supervision},
  author  = {Ratner, Alexander and Bach, Stephen H. and Ehrenberg, Henry and Fries, Jason and Wu, Sen and R{\'e}, Christopher},
  journal = {The VLDB Journal},
  volume  = {29},
  number  = {2--3},
  pages   = {709--730},
  year    = {2020},
  doi     = {10.1007/s00778-019-00552-1},
  url     = {https://doi.org/10.1007/s00778-019-00552-1}
}

@article{lamsal2021covidsentiment,
  title   = {Design and analysis of a large-scale {COVID-19} tweets dataset},
  author  = {Lamsal, Rabindra},
  journal = {Applied Intelligence},
  volume  = {51},
  number  = {5},
  pages   = {2790--2804},
  year    = {2021},
  doi     = {10.1007/s10489-020-02029-z},
  url     = {https://doi.org/10.1007/s10489-020-02029-z}
}

@inproceedings{wu2025mabsa,
  title     = {{M}-{ABSA}: A Multilingual Dataset for Aspect-Based Sentiment Analysis},
  author    = {Wu, ChengYan and Ma, Bolei and Liu, Yihong and Zhang, Zheyu and Deng, Ningyuan and Li, Yanshu and Chen, Baolan and Zhang, Yi and Xue, Yun and Plank, Barbara},
  booktitle = {Proceedings of the 2025 Conference on Empirical Methods in Natural Language Processing},
  month     = nov,
  year      = {2025},
  address   = {Suzhou, China},
  publisher = {Association for Computational Linguistics},
  pages     = {2530--2557},
  doi       = {10.18653/v1/2025.emnlp-main.128},
  url       = {https://aclanthology.org/2025.emnlp-main.128/}
}

@article{zhang2022absa,
  title   = {A Survey on Aspect-Based Sentiment Analysis: Tasks, Methods, and Challenges},
  author  = {Zhang, Wenxuan and Li, Xin and Deng, Yang and Bing, Lidong and Lam, Wai},
  journal = {IEEE Transactions on Knowledge and Data Engineering},
  volume  = {35},
  number  = {11},
  pages   = {11019--11038},
  year    = {2023},
  doi     = {10.1109/TKDE.2022.3230975},
  url     = {https://doi.org/10.1109/TKDE.2022.3230975}
}

@article{schroeder2016bayesact,
  title     = {Modeling Dynamic Identities and Uncertainty in Social Interactions: {B}ayesian Affect Control Theory},
  author    = {Schr{\"o}der, Tobias and Hoey, Jesse and Rogers, Kimberly B.},
  journal   = {American Sociological Review},
  volume    = {81},
  number    = {4},
  pages     = {828--855},
  year      = {2016},
  doi       = {10.1177/0003122416650963},
  url       = {https://doi.org/10.1177/0003122416650963}
}

@inproceedings{matsuda2025dependency,
  title     = {Step-by-step Instructions and a Simple Tabular Output Format Improve the Dependency Parsing Accuracy of {LLM}s},
  author    = {Matsuda, Hiroshi and Ma, Chunpeng and Asahara, Masayuki},
  booktitle = {Proceedings of the 18th International Conference on Parsing Technologies (IWPT, SyntaxFest 2025)},
  month     = aug,
  year      = {2025},
  address   = {Ljubljana, Slovenia},
  publisher = {Association for Computational Linguistics},
  pages     = {11--19},
  url       = {https://aclanthology.org/2025.iwpt-1.2/}
}

@article{zorenbohmer2026emograce,
  title   = {{EmoGRACE}: aspect-based emotion analysis for social media data},
  author  = {Zorenb{\"o}hmer, Christina and Schmidt, Sebastian and Hanny, David and Resch, Bernd},
  journal = {Social Network Analysis and Mining},
  volume  = {16},
  pages   = {42},
  year    = {2026},
  doi     = {10.1007/s13278-026-01585-5},
  url     = {https://doi.org/10.1007/s13278-026-01585-5}
}

\FloatBarrier
\clearpage

\appendix
\begin{center}
    {\large Appendix\par}
\end{center}

\section{Task Logic and Example Annotations}

\subsection{Task Logic Visualization}

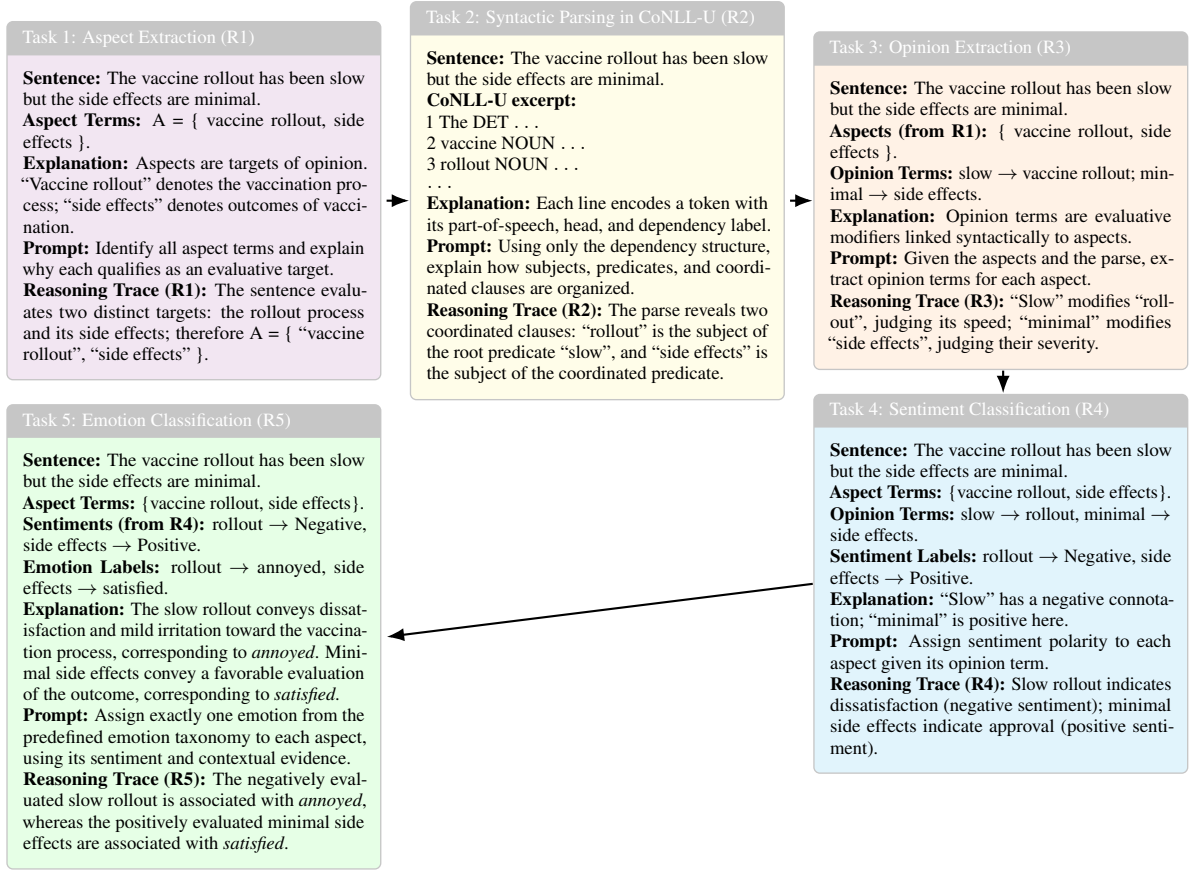
\begin{figure*}[!t]
\centering
\centering
\scriptsize
\setlength{\fboxsep}{0pt}
\newlength{\synboxwidth}
\setlength{\synboxwidth}{0.31\textwidth}

\begin{tikzpicture}[
  node distance=3mm and 3mm,
  >=Latex,
  every node/.style={inner sep=0pt},
  boxnode/.style={align=left},
  flow/.style={->, thick}
]

\node (t2) [boxnode] {
\begin{tcolorbox}[
  enhanced,
  title={Task 1: Aspect Extraction (R1)},
  colback=violet!10,
  colframe=gray!45,
  boxrule=0.5pt,
  arc=2pt,
  left=3pt,
  right=3pt,
  top=3pt,
  bottom=3pt,
  width=\synboxwidth
]
\textbf{Sentence:} The vaccine rollout has been slow but the side effects are minimal.

\textbf{Aspect Terms:} A = \{ vaccine rollout, side effects \}.

\textbf{Explanation:}
Aspects are targets of opinion. ``Vaccine rollout'' denotes the
vaccination process; ``side effects'' denotes outcomes of vaccination.

\textbf{Prompt:}
Identify all aspect terms and explain why each qualifies as an evaluative target.

\textbf{Reasoning Trace (R1):}
The sentence evaluates two distinct targets: the rollout process and its
side effects; therefore A = \{ ``vaccine rollout'', ``side effects'' \}.
\end{tcolorbox}
};

\node (t1) [right=3mm of t2, boxnode] {
\begin{tcolorbox}[
  enhanced,
  title={Task 2: Syntactic Parsing in CoNLL-U (R2)},
  colback=yellow!10,
  colframe=gray!45,
  boxrule=0.5pt,
  arc=2pt,
  left=3pt,
  right=3pt,
  top=3pt,
  bottom=3pt,
  width=\synboxwidth
]
\textbf{Sentence:} The vaccine rollout has been slow but the side effects are minimal.

\textbf{CoNLL-U excerpt:}\\
1 The DET $\dots$\\
2 vaccine NOUN $\dots$\\
3 rollout NOUN $\dots$\\[-0.4ex]
$\ldots$

\textbf{Explanation:}
Each line encodes a token with its part-of-speech, head, and dependency label.

\textbf{Prompt:}
Using only the dependency structure, explain how subjects, predicates,
and coordinated clauses are organized.

\textbf{Reasoning Trace (R2):}
The parse reveals two coordinated clauses: ``rollout'' is the subject of
the root predicate ``slow'', and ``side effects'' is the subject of the
coordinated predicate.
\end{tcolorbox}
};

\node (t3) [right=3mm of t1, boxnode] {
\begin{tcolorbox}[
  enhanced,
  title={Task 3: Opinion Extraction (R3)},
  colback=orange!10,
  colframe=gray!45,
  boxrule=0.5pt,
  arc=2pt,
  left=3pt,
  right=3pt,
  top=3pt,
  bottom=3pt,
  width=\synboxwidth
]
\textbf{Sentence:} The vaccine rollout has been slow but the side effects are minimal.

\textbf{Aspects (from R1):} \{ vaccine rollout, side effects \}.

\textbf{Opinion Terms:}
slow $\rightarrow$ vaccine rollout; minimal $\rightarrow$ side effects.

\textbf{Explanation:}
Opinion terms are evaluative modifiers linked syntactically to aspects.

\textbf{Prompt:}
Given the aspects and the parse, extract opinion terms for each aspect.

\textbf{Reasoning Trace (R3):}
``Slow'' modifies ``rollout'', judging its speed; ``minimal'' modifies
``side effects'', judging their severity.
\end{tcolorbox}
};

\node (t5) [below=3mm of t2, boxnode] {
\begin{tcolorbox}[
  enhanced,
  title={Task 5: Emotion Classification (R5)},
  colback=green!10,
  colframe=gray!45,
  boxrule=0.5pt,
  arc=2pt,
  left=3pt,
  right=3pt,
  top=3pt,
  bottom=3pt,
  width=\synboxwidth
]
\textbf{Sentence:} The vaccine rollout has been slow but the side effects are minimal.

\textbf{Aspect Terms:} \{vaccine rollout, side effects\}.\\
\textbf{Sentiments (from R4):}
rollout $\rightarrow$ Negative,
side effects $\rightarrow$ Positive.

\textbf{Emotion Labels:}
rollout $\rightarrow$ annoyed,
side effects $\rightarrow$ satisfied.

\textbf{Explanation:}
The slow rollout conveys dissatisfaction and mild irritation toward the
vaccination process, corresponding to \emph{annoyed}. Minimal side
effects convey a favorable evaluation of the outcome, corresponding to
\emph{satisfied}.

\textbf{Prompt:}
Assign exactly one emotion from the predefined emotion taxonomy to each
aspect, using its sentiment and contextual evidence.

\textbf{Reasoning Trace (R5):}
The negatively evaluated slow rollout is associated with
\emph{annoyed}, whereas the positively evaluated minimal side effects
are associated with \emph{satisfied}.
\end{tcolorbox}
};

\node (t4) [below=3mm of t3, boxnode] {
\begin{tcolorbox}[
  enhanced,
  title={Task 4: Sentiment Classification (R4)},
  colback=cyan!10,
  colframe=gray!45,
  boxrule=0.5pt,
  arc=2pt,
  left=3pt,
  right=3pt,
  top=3pt,
  bottom=3pt,
  width=\synboxwidth
]
\textbf{Sentence:} The vaccine rollout has been slow but the side effects are minimal.

\textbf{Aspect Terms:} \{vaccine rollout, side effects\}.\\
\textbf{Opinion Terms:} slow $\rightarrow$ rollout, minimal $\rightarrow$ side effects.

\textbf{Sentiment Labels:}
rollout $\rightarrow$ Negative, side effects $\rightarrow$ Positive.

\textbf{Explanation:}
``Slow'' has a negative connotation; ``minimal'' is positive here.

\textbf{Prompt:}
Assign sentiment polarity to each aspect given its opinion term.

\textbf{Reasoning Trace (R4):}
Slow rollout indicates dissatisfaction (negative sentiment); minimal
side effects indicate approval (positive sentiment).
\end{tcolorbox}
};

\draw[flow] (t2.east) -- (t1.west);
\draw[flow] (t1.east) -- (t3.west);
\draw[flow] (t3.south) -- (t4.north);
\draw[flow] (t4.west) -- (t5.east);

\end{tikzpicture}
\caption{Reasoning example for aspect-based sentiment analysis and emotion detection on a single sentence via our teaching framework.}
\label{fig:pipeline}
\end{figure*}

Figure~\ref{fig:pipeline} presents the reasoning structure of the proposed framework.

Given a raw tweet, the model proceeds through five consecutive tasks: aspect extraction, syntactic parsing, opinion extraction, sentiment classification, and emotion classification. This decomposition follows prior work in aspect-based sentiment analysis and structured reasoning \cite{pontiki2014semeval,zhang2022absa,fan2025synchain}, enabling the model to identify both targets and their associated sentiment and emotion.

The visualization emphasizes pipeline hierarchy. Aspect extraction determines the opinion targets, syntactic parsing provides grammatical structure, and opinion extraction links expressions to aspects. These intermediate representations then support aspect-level sentiment and emotion prediction. In our teacher--student setting, the teacher produces both labels and reasoning traces for each step, and the student is trained to imitate this full process, similar to recent structured reasoning approaches in ABSA \cite{fan2025synchain}.

\subsection{Example Annotations}

To illustrate the complete reasoning pipeline, we present representative examples that demonstrate our model's capability to handle diverse COVID-19 discourse. These examples showcase the five-task reasoning chain (R1--R5) and highlight how the model disambiguates sentiment and emotion at the aspect level.

\subsubsection{Handling Context}

\begin{figure*}[!t]
\centering
\input{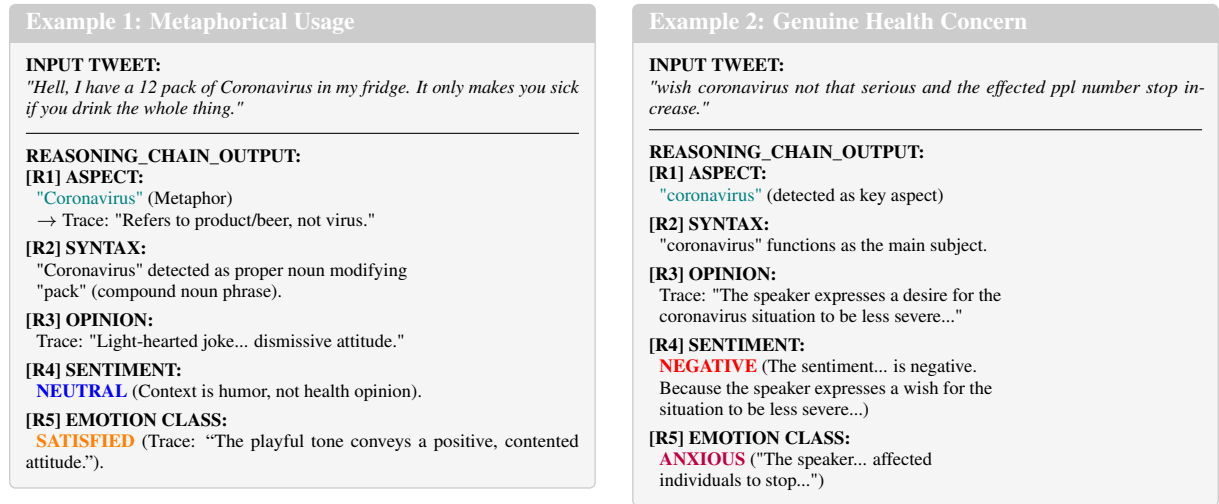}
\caption{Example reasoning traces for two COVID-19 tweets demonstrating the complete annotation pipeline. Left: sarcastic reference requiring contextual sentiment disambiguation. Right: anxious expression of genuine health concern with negative sentiment.}
\label{fig:reasoning_examples}
\end{figure*}

Figure~\ref{fig:reasoning_examples} presents two contrasting cases that demonstrate the importance of aspect-level analysis.

The first example illustrates the model's ability to correctly identify metaphorical usage of "Coronavirus" (referring to Corona beer rather than the virus), resulting in neutral sentiment and satisfied emotion despite surface-level ambiguity. The syntactic parsing (R2) and opinion extraction (R3) steps are crucial here, as they reveal the humorous context through dependency relationships.

In contrast, the second example shows genuine health concern, where incomplete grammar and informal language ("wish coronavirus not that serious") are correctly interpreted as expressing negative sentiment and anxiety toward the pandemic situation. These examples demonstrate how the reasoning traces enable the model to navigate linguistic complexity and contextual meaning that would challenge traditional sentiment analysis approaches.

\subsubsection{Handling Different Emotions in Multi-Aspect Example}

\begin{figure*}[!t]
\centering
\includegraphics[width=0.55\textwidth]{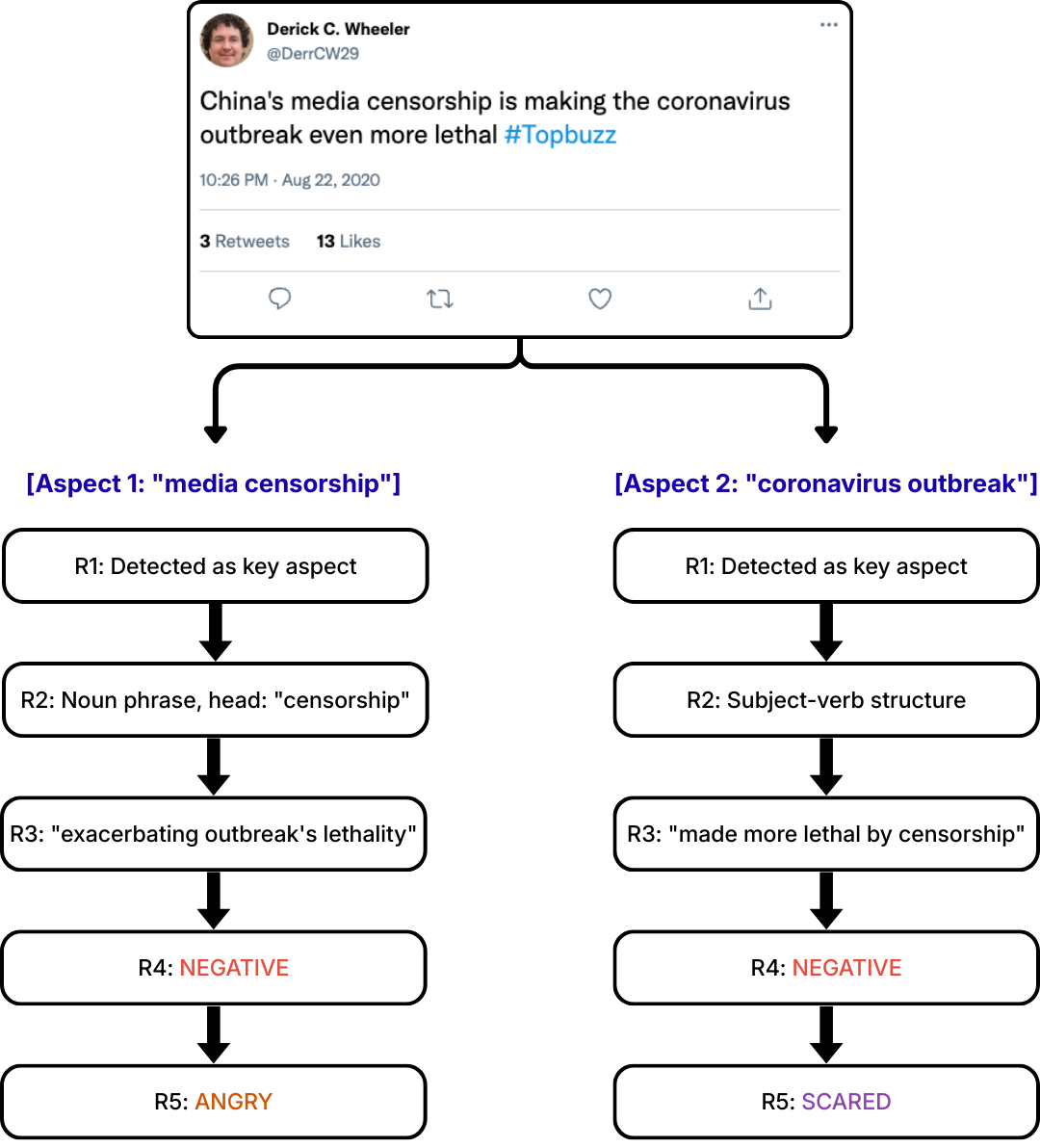}
\caption{Illustrative reconstruction of a real tweet. Display name, profile image, username, timestamp, and engagement metadata are fictionalized. Reasoning trace across multiple aspects for a COVID-19 tweet. The model identifies two distinct aspects and performs the complete annotation pipeline for each, resulting in shared negative sentiment but divergent emotions: \textit{angry} toward media censorship versus \textit{scared} regarding the outbreak itself.}
\label{fig:multiaspect_example}
\end{figure*}

Figure~\ref{fig:multiaspect_example} demonstrates the model's capability on a more complex tweet containing two distinct evaluative targets.

The model successfully identifies both "media censorship" and "coronavirus outbreak" as separate aspects, performing independent reasoning chains for each. While both aspects receive negative sentiment classifications (R4), the emotion labels (R5) differ: anger toward censorship policies versus fear regarding outbreak severity. This distinction is semantically meaningful since anger reflects frustration with human decisions (censorship), while fear responds to the health threat itself.

The ability to extract and independently analyze multiple aspects within a single tweet is essential for understanding nuanced public discourse during crisis situations, where opinions often target multiple entities simultaneously with varying emotional valences.

\subsubsection{Summary}

These examples collectively demonstrate our model's capabilities: handling informal social media language, disambiguating metaphorical and literal usage, attributing sentiments and identifying fine-grained emotions to individual aspects.

\section{Emotional and Sentiment Analysis of Teacher Annotations}

\subsection{Dataset Preprocessing}
\label{app:preprocessing}

\paragraph{News Filtering}
A challenge using these datasets for aspect-based sentiment analysis is the number of ``news headline'' style tweets, such as factual statements, news links, and quoted headlines that lack personal opinion or emotion, and therefore introduce irrelevant noise during training. Such tweets are unsuitable for ABSA, as they do not express aspect-level sentiment. Tweets identified as news were removed during preprocessing. News tweets were detected using patterns such as URLs, common headline-style prefixes, source attributions, and truncated headline snippets.

\paragraph{Text Preprocessing}
The remaining tweets underwent additional preprocessing, including emoji removal, Unicode normalization using the \texttt{ftfy}~\cite{speer2019ftfy} library, and URL stripping. We then performed syntactic parsing using spaCy's~\cite{honnibal2020spacy} \texttt{en\_core\_web\_sm} model, converting each tweet to CoNLL-U format, which encodes token-level information such as lemma, part-of-speech tags, and dependency relations.

\subsection{Dataset Statistics}
\label{app:dataset_statistics}

We report statistics computed from the
Qwen2.5-32B~\cite{qwen2_5} teacher annotations, aggregated over the
COVID19NLP~\cite{covid19nlp} and
COVIDSenti~\cite{naseem2021covidsenti} datasets.

The resulting corpus contains 21{,}937 sentences and 50{,}615 aspect
instances, corresponding to an average of 2.307 aspects per sentence.
During aggregation, we detected 13 orphan sentiment labels and 13 orphan
emotion labels, defined as entries with missing aspect identifiers in the
teacher model's JSON output. These entries correspond to 0.026\% of the
aspect instances and were excluded from the training data and all
subsequent analyses.

\begin{table}[t]
\centering
\small
\caption{Corpus summary based on Qwen2.5-32B annotations.}
\label{tab:appendix_corpus_summary}
\begin{tabular}{lr}
\toprule
\textbf{Metric} & \textbf{Value} \\
\midrule
Total sentences & 21,937 \\
Total aspects & 50,615 \\
Average aspects per sentence & 2.307 \\
\bottomrule
\end{tabular}
\end{table}

\begin{table}[t]
\centering
\small
\caption{Aspect-level sentiment distribution
($N=50{,}615$).}
\label{tab:asp_level_summary}
\begin{tabular}{lrr}
\toprule
\textbf{Sentiment} & \textbf{Count} & \textbf{Percent} \\
\midrule
Neutral  & 26,249 & 51.87\% \\
Negative & 18,288 & 36.13\% \\
Positive & 6,078  & 12.01\% \\
\bottomrule
\end{tabular}
\end{table}

\begin{table}[t]
\centering
\small
\caption{Aspect-level emotion distribution
($N=50{,}615$). The \textit{no emotion} label indicates that the teacher
did not assign an emotion to the aspect.}
\label{tab:emotion_dist}
\begin{tabular}{lrr}
\toprule
\textbf{Emotion} & \textbf{Count} & \textbf{Percent} \\
\midrule
Neutral      & 21,608 & 42.70\% \\
Annoyed      & 10,488 & 20.73\% \\
Anxious      & 8,877  & 17.54\% \\
Hopeful      & 1,640  & 3.24\% \\
Scared       & 1,552  & 3.07\% \\
Empathetic   & 1,436  & 2.84\% \\
Satisfied    & 1,017  & 2.01\% \\
Angry        & 1,014  & 2.00\% \\
Optimistic   & 818    & 1.62\% \\
Pessimistic  & 684    & 1.35\% \\
Trustful     & 550    & 1.09\% \\
Sad          & 449    & 0.89\% \\
Thankful     & 262    & 0.52\% \\
Proud        & 150    & 0.30\% \\
No emotion   & 70     & 0.14\% \\
\bottomrule
\end{tabular}
\end{table}

We analyze the distribution of emotion annotations across aspect categories to better understand the structure of public discourse captured by our teacher model.

To improve interpretability, semantically equivalent aspect mentions (e.g., \textit{coronavirus}, \textit{covid}, \textit{corona pandemic}, \textit{covid crisis}) are normalized into unified aspect categories.

\begin{figure*}[!t]
\centering
\includegraphics[width=\textwidth]{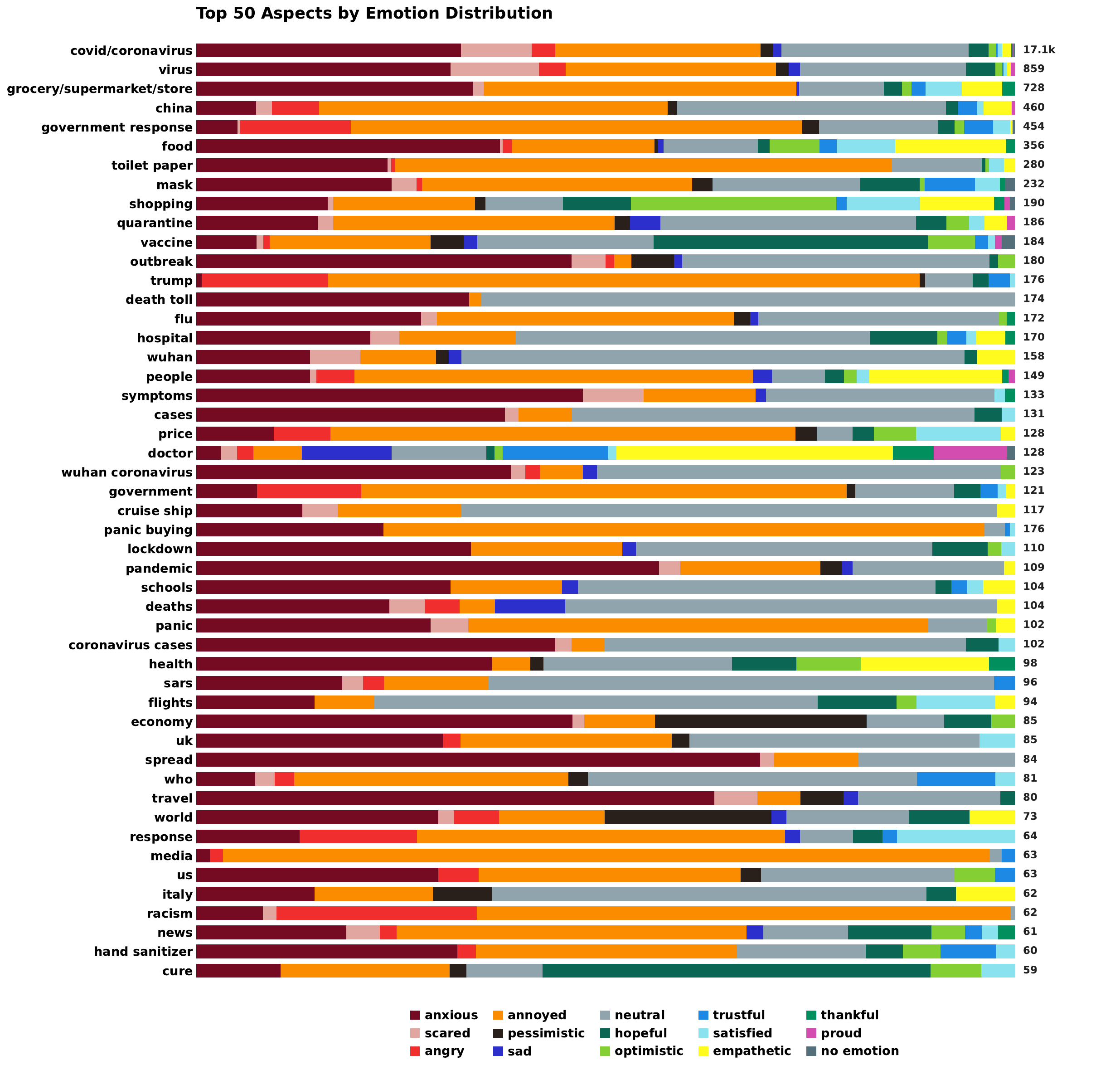}
\caption{Emotion distribution across the 50 most frequent aspects. Each bar represents the normalized distribution of emotion categories for a given aspect.}
\label{fig:top50_emotions}
\end{figure*}

Figure~\ref{fig:top50_emotions} presents the emotion distributions for the 50 most frequent aspects. The results show that aspects related to pandemic are primarily associated with negative emotions, particularly anxiety and annoyance. In contrast, consumer-related aspects (e.g., \textit{shopping}, \textit{food}) exhibit more diverse emotional profiles.

The \textit{vaccine} aspect is associated with hopeful emotion, reflecting expectations of recovery. In contrast, aspects related to political entities (e.g., \textit{Trump}, \textit{government response}) are largely associated with annoyance, indicating frustration toward decision-making and policy handling.

Overall, negative emotional valence dominates the dataset, with anxiety and annoyance being the most prevalent emotions, as public discourse during the pandemic is characterized by uncertainty and frustration. Aspects related to the pandemic often evoke anxiety due to perceived health risks and uncertainty, while secondary effects such as price increases, supply shortages, and quarantine measures contribute to annoyance and anger. Due to the large number of neutral tweets, where users write summaries or reports regarding certain aspects, neutral emotions account for a substantial proportion of each aspect’s emotion distribution.

\section{Human Annotation Set}

The gold standard dataset was annotated by four volunteer annotators from the research team, without financial compensation. Annotators were trained using a coding scheme defining emotion and sentiment categories, examples, and rules for aspect spans, sarcasm, and edge cases. A calibration session was conducted before independent annotation.

In total, annotators labeled 500 tweets with intentionally overlapping assignments to enable agreement analysis. Annotator 1 labeled tweets 1--250, Annotator 2 labeled tweets 250--500, Annotator 3 labeled tweets 1--125 and 375--500, and Annotator 4 labeled tweets 125--375.

\begin{comment}
    In Figure~\ref{fig:annotation_tool}, we show our in-house annotation tool, which supports aspect-based sentiment and emotion labeling.

\begin{figure}[t]
    \centering
    \includegraphics[width=\linewidth]{figures/annotation_tool.png}
    \caption{Our in-house annotation tool for aspect-based sentiment and emotion labeling.}
    \label{fig:annotation_tool}
\end{figure}
\end{comment}

This design ensures that each subset of tweets is annotated by at least two annotators, allowing us to examine inter-coding agreement, following standard practices in dataset construction for sentiment and emotion analysis \cite{demszky2020goemotions,naseem2021covidsenti}.

We report aspect agreement in F1 score, sentiment accuracy, and emotion accuracy between overlapping annotator pairs in Table~\ref{tab:annotation_agreement}.

\begin{table}[t]
\centering
\small
\begin{tabular}{lccc}
\hline
\textbf{Pair} & \textbf{Aspect (F1)} & \textbf{Sent.} & \textbf{Emo.} \\
\hline
A1--A3 & 0.7493 & 0.8635 & 0.8229 \\
A1--A4 & 0.7147 & 0.8289 & 0.7909 \\
A2--A4 & 0.7386 & 0.7676 & 0.7042 \\
A2--A3 & 0.7101 & 0.8516 & 0.8438 \\
\hline
\textbf{Mean} & 0.7282 & 0.8279 & 0.7905 \\
\hline
\end{tabular}
\caption{Inter-annotator agreement (aspect F1, sentiment accuracy, and emotion agreement).}
\label{tab:annotation_agreement}
\end{table}

\end{document}